\documentclass{article}

     \usepackage[preprint]{neurips_2021}

\usepackage[utf8]{inputenc} 
\usepackage[T1]{fontenc}    
\usepackage{hyperref}       
\usepackage{url}            
\usepackage{booktabs}       
\usepackage{amsfonts}       
\usepackage{nicefrac}       
\usepackage{microtype}      
\usepackage{xcolor}         

\usepackage{graphicx}
\usepackage{algorithm}
\usepackage{algorithmic}
\usepackage{subcaption}
\usepackage{multirow}
\usepackage{amsmath}
\usepackage{array}

\graphicspath{{figures/}}

\title{Dual GNNs: Graph Neural Network Learning with Limited Supervision}

\author{
Abdullah Alchihabi,  Yuhong Guo\\
Carleton University, Ottawa, Canada\\
\texttt{abdullahalchihabi@cmail.carleton.ca, yuhong.guo@carleton.ca}
}

\begin{document}

\maketitle

\begin{abstract}
Graph Neural Networks (GNNs) require a relatively large number of labeled nodes and a reliable/uncorrupted graph connectivity structure
	in order to obtain good performance on the semi-supervised node classification task. The performance of GNNs can degrade significantly as the number of 
labeled nodes
	decreases or the graph connectivity structure 
	is corrupted by adversarial attacks or due to noises in data measurement/collection. 
	Therefore, it is important to develop GNN models 
	that are able to achieve good performance when 
	there is limited supervision knowledge -- a few labeled nodes and noisy graph structures.
	In this paper, we propose a novel Dual GNN learning framework to address this challenge task. 
	The proposed framework has two GNN based node prediction modules. 
	The primary 
	module uses the input graph structure to 
	induce regular node embeddings and predictions with a regular GNN baseline,
	while the auxiliary 
	module constructs a new graph structure through fine-grained spectral clusterings
	and learns new node embeddings and predictions. 
	By integrating the two modules in a dual GNN learning framework, we perform joint learning 
	in an end-to-end fashion. 
	This general framework can be applied on many GNN baseline models. 
	The experimental results validate that the proposed dual GNN framework can greatly outperform the GNN baseline methods
	when the labeled nodes are scarce and the graph connectivity structure is noisy. 

\end{abstract}

\section{Introduction}

\begin{figure}[t]
\begin{subfigure}{0.5\textwidth}
\centering	
\includegraphics[width= 0.52\textwidth]{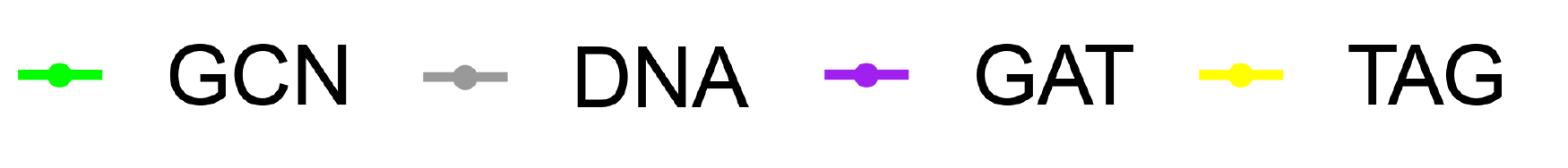}
{\includegraphics[width=  .8\textwidth,height=1.4in]{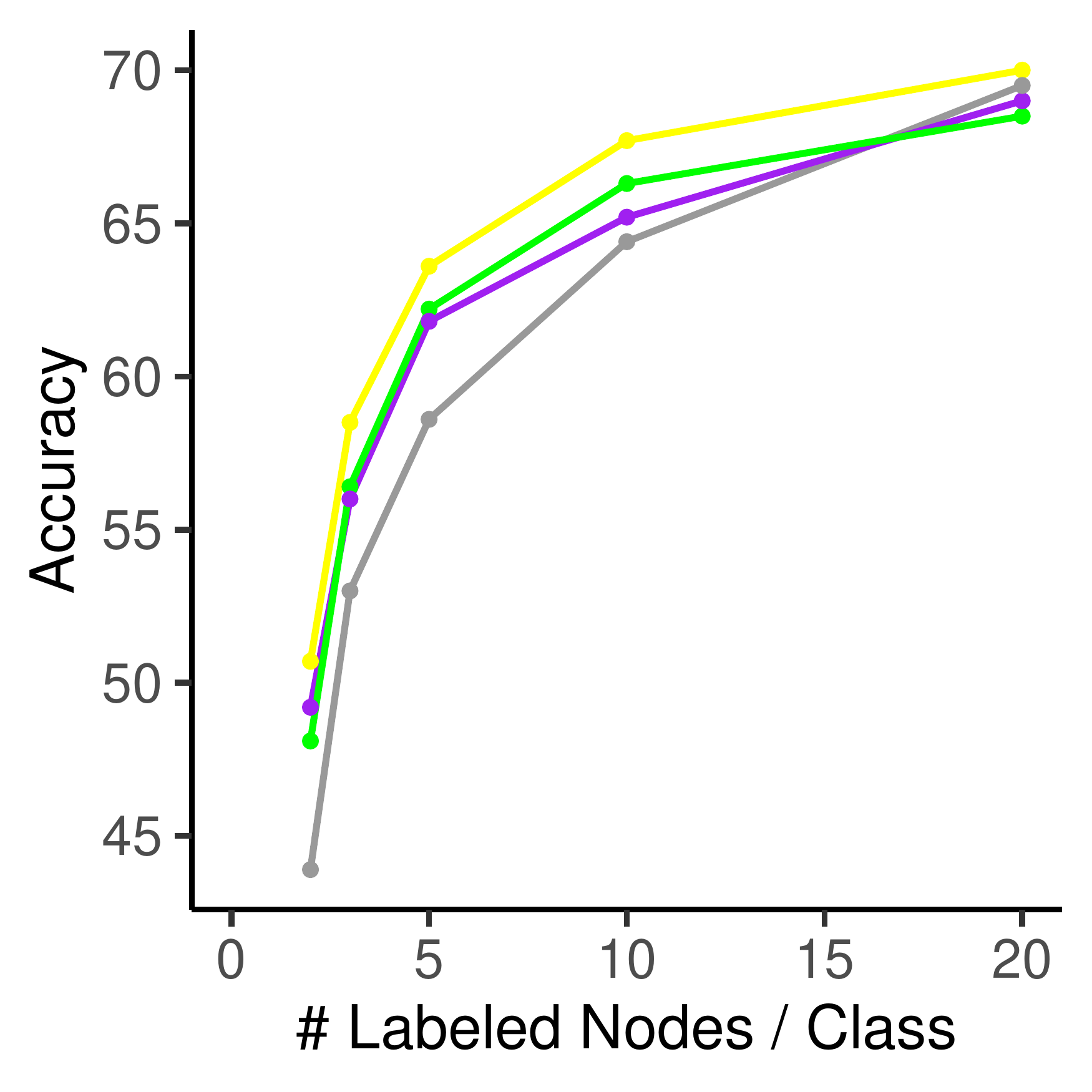}} 
\caption{Few Labeled Nodes}
\label{fig:1_a}
\end{subfigure}
\begin{subfigure}{0.5\textwidth}
\centering	
\includegraphics[width= 0.52\textwidth]{Legend_CiteSeer_Baselines.pdf}
{\includegraphics[width = .8 \textwidth,height=1.4in]{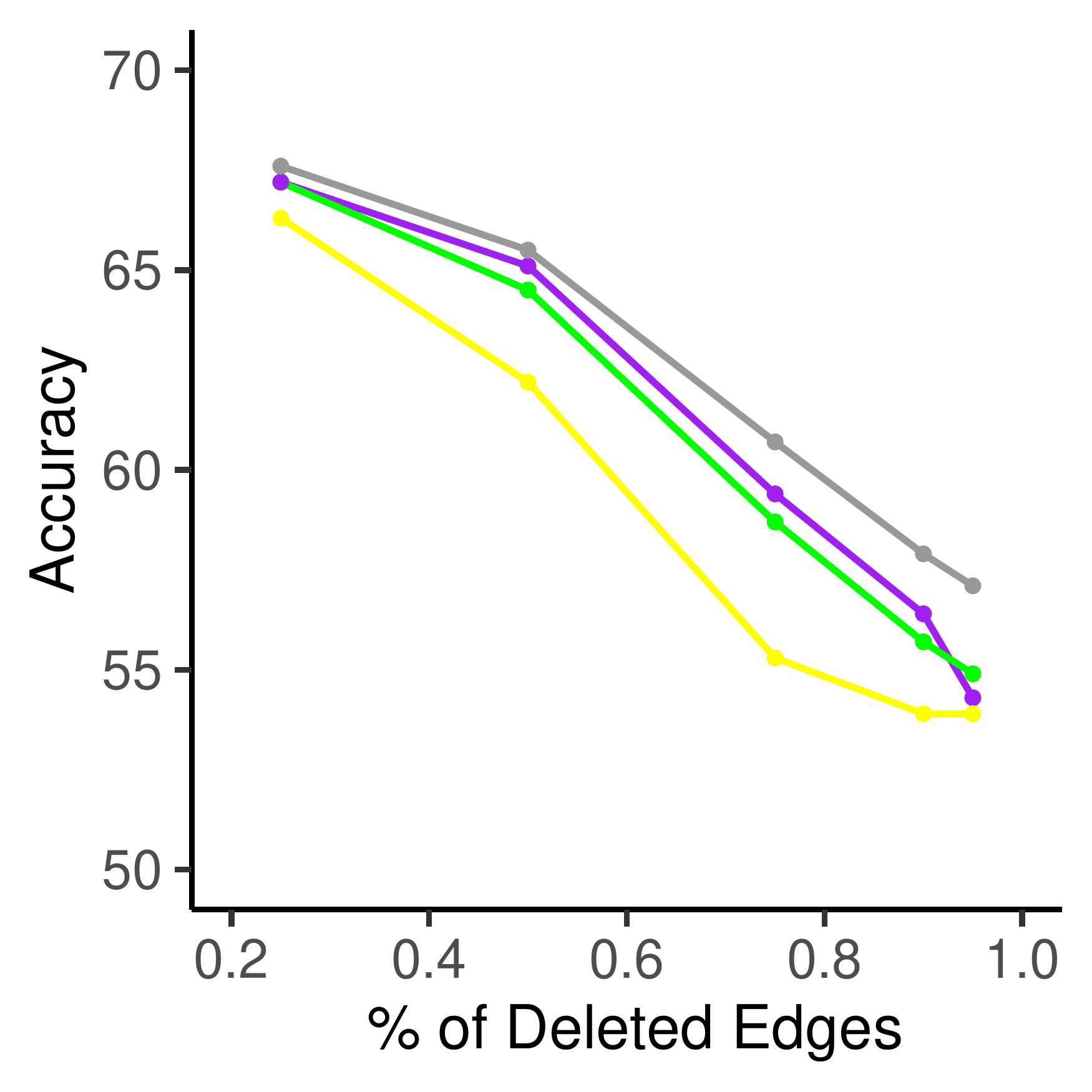}}
\caption{Noisy Graph Structure}
\label{fig:1_b}
\end{subfigure}
\caption{Performance degradation of GNN models on CiteSeer: (a) Few Labeled Nodes; (b) Noisy Graph Structures.}
\end{figure}

Graph Neural Networks (GNN) have been successfully employed to solve multiple tasks such as node classification, graph completion, and edge prediction across a variety of application domains including computational chemistry \cite{shi2020graph}, protein-protein interactions \cite{zitnik2017predicting} and knowledge-base completion \cite{schlichtkrull2018modeling}. 
In particular, many GNN advancements have addressed the typical node classification task 
in a semi-supervised learning setting where only a subset of nodes in the graph are labeled,
including the well known 
Graph Convolutional Networks (GCNs) \cite{kipf2016semi}, 
Graph Attention Networks (GATs) \cite{velivckovic2017graph}, 
Topology Adaptive Graph Convolutional Networks (TAGs) \cite{du2017topology} 
and Dynamic Neighborhood Aggregation networks (DNAs) \cite{fey2019just}. 
These GNN models 
have achieved great results on the benchmark GNN learning datasets. 
However, they typically require a large number of labeled nodes as well as 
reliable/uncorrupted graph structures 
in order to obtain good performance. 
Their performance can degrade significantly as the number of labeled nodes becomes scarce \cite{li2018deeper,lin2020shoestring} (as shown in Figure \ref{fig:1_a}) or when the graph structures are noisy or corrupted \cite{wang2020graph,chen2020iterative} 
such as many edges are deleted (as shown in Figure \ref{fig:1_b}).
%

The performance drop of GNNs with limited labeled data can be explained by the inability of GNNs to 
propagate the label information 
from the labeled nodes to the rest of the graph. 
That is, while it is known that deep GNN architectures can cause over-smoothing problems~\cite{li2018deeper},
a relatively shallow GNN architecture can fail to propagate messages across the whole graph
and cause the classifier to overfit the 
small neighborhoods of the labeled nodes. 
When there are very small numbers of labeled nodes,
this will induce serious overfitting problems and degrade the classification performance \cite{li2018deeper,sun2019multi,lin2020shoestring}. 
In addition, GNNs learn discriminative node embeddings for effective node classification
by propagating messages across the edges of the graph.
Hence noisy or corrupted graph structures can greatly impair the message passing process
and degrade the ultimate node classification performance
\cite{li2018deeper,wang2020graph}. 
On the other hand, 
the dependence of GNN models on a large number of labeled nodes and reliable graph structures 
can seriously limit their applications, 
as in many application domains it is very expensive or difficult to obtain a large number of labeled examples. 
It is also very difficult to guarantee reliable/uncorrupted graph structures given the numerous sources of noises 
that could damage the graph connectivity structures, ranging from adversarial attacks on the graph structures to 
data collection or measurement noises 
\cite{wang2020graph,chen2020iterative}.  
Therefore, it is important to develop GNN models that can learn efficiently with few labeled nodes and are robust to corrupted/unreliable graph structures.

In this work, we propose a novel dual GNN learning framework that is resilient to low label rates 
and corrupted/noisy graph structures with deleted edges. 
The proposed framework is made up of two node prediction modules. 
The first module can be treated as
a standard primary 
GNN model which takes the original graph data as input. 
It suffers from the aforementioned message propagation drawbacks 
when labeled nodes are scarce and graph structures are noisy.
To address this problem, the second GNN 
module employs a fine-grained spectral clustering method based on the 
node embedding results of the first module 
to construct a new adjacency matrix and hence a new graph structure,
aiming to enable effective information propagation across the graph, 
and facilitate the subsequent node embedding and node classification learning. 
The two modules coordinate with each other within the integrated dual learning framework 
to perform end-to-end training with a joint objective function. 
This general dual learning framework can be applied on many standard GNN baseline models. 
We conduct experiments with four baseline graph neural network learning models: 
GCNs \cite{kipf2016semi}, GATs \cite{velivckovic2017graph}, TAGs \cite{du2017topology}, and DNAs \cite{fey2019just}. 
The experimental results show that the 
proposed framework can significantly improve the baseline models
when the labeled nodes are very scarce and graph structure is noisy/corrupted across multiple benchmark datasets.

\section{Related Works}

Graph neural network (GNN) learning has received a lot of attention in the research community.
Many early works developed
spectral approaches for GNN learning based on graph Laplacians 
\cite{hammond2011wavelets,bruna2013spectral,henaff2015deep,defferrard2016convolutional,kipf2016semi}. 
For example, Defferrard et al. employed Chebyshev polynomial to parameterize the learned filters and localize them in the spectral domain \cite{defferrard2016convolutional}. 
Kipf et al. proposed a Graph Convolutional Network (GCN) model 
that uses a first-order approximation of the Chebyshev polynomial with a re-normalization trick to ensure numerical stability during training \cite{kipf2016semi}. 
Another set of works 
operate directly in the spatial domain by defining novel message-passing and message-aggregation functions 
\cite{velivckovic2017graph,du2017topology,fey2019just}. 
Velivckovic et al. proposed a Graph Attention Network (GAT) 
that uses a self-attention mechanism to assign importance weights to 
the edges of graphs
\cite{velivckovic2017graph}. Du et al. proposed a Topology Adaptive Graph Convolutional Network (TAG) 
that aggregates messages from different powers of the adjacency matrix 
to enlarge the receptive field of the nodes \cite{du2017topology}. 
Fey proposed a Dynamic Neighborhood Aggregation (DNA) model
that defines dynamic receptive fields for each node 
by allowing some nodes to aggregate global information from the graph while other nodes focus on local information \cite{fey2019just}.
Most of these previous works have focused on developing powerful GNNs with a greater representation power to solve challenging graph-related tasks, including node classification. 

More recently, 
a number of methods have been proposed to address some notable 
limitations for standard GNN models, 
including
the vulnerability to adversarial attacks \cite{elinas2019variational,you2020does,zhang2020gnnguard,geisler2020reliable,wang2020graph}, 
over-smoothing \cite{li2018deeper,rong2020dropedge,min2020scattering}, 
sensitivity to label noise \cite{de2020analysis}, and performance degradation 
with scarce labels \cite{sun2019multi,zhou2019dynamic,wang2020graph,calder2020poisson,lin2020shoestring} or 
data corruptions \cite{you2020handling}.
In particular, the label scarcity problem has been addressed by using various strategies, including
self-supervision \cite{sun2019multi}, self-training \cite{zhou2019dynamic} and metric learning \cite{lin2020shoestring}. 
Lin et al. proposed a new framework that employs metric learning to improve the performance of GNNs under very low label rates \cite{lin2020shoestring}. Zhou et al. used dynamic self-training to increase the size of the training set and alleviate overfitting problems when the labels are scarce \cite{zhou2019dynamic}. Sun et al. used self-supervision and self-training to augment the training set with confidently labeled nodes whose self-supervised labels match their predicted pseudo-labels in order to increase the number of labeled samples \cite{sun2019multi}.  

In addition, GNNs have been shown to be vulnerable to a variety of adversarial attacks. 
A number of works have proposed algorithms to defend against adversarial attacks on GNNs. 
You et al. employed multiple self-supervised tasks with adversarial training to improve the robustness of GNNs to single-node evasion attacks \cite{you2020does}. Abu-El-Haija et al. proposed a GNN architecture that trains multiple GCNs on nodes sampled using random walks to induce tolerance to node feature perturbations \cite{abu2020n}. Chen et al. proposed an iterative deep graph learning framework to defend against
adversarial attacks on graph structures \cite{chen2020iterative}. 
Geisler et al. presented a novel message aggregation function that is resilient to adversarial attacks on graph structures \cite{geisler2020reliable}. Elinas et al. introduced a framework that uses variational inference 
to defend against adversarial attacks on graph structures \cite{elinas2019variational}. 
Wang et al. addressed the issues of noisy graph structures and few labeled nodes by introducing a framework to learn a family of classification functions using amortised variational inference \cite{wang2020graph}. 
A comprehensive review of adversarial attacks and defense methods on GNNs 
is provided in \cite{sun2018adversarial}.

\section{Method}

\subsection{Problem Setup}
We consider the following transductive semi-supervised node classification setting.
The input 
is a graph $G = (V,E)$, where $V$ is the set of nodes with size $|V| = N$ and $E$ is the set of edges. 
$E$ is typically represented by an adjacency matrix $A$ of size $N\times N$,
which can be either symmetric (for undirected graphs) or asymmetric (for directed graphs),
either have weights or binary values. 

Each node in the graph $G$ is associated with a corresponding feature vector of size $D$. 
The feature vectors of all the nodes in the graph are represented by an input feature matrix $X\in\mathbb{R}^{N\times D}$. 
In the context of transductive semi-supervised node classification, 
the nodes in $V$ are split into two subsets: a subset of labeled nodes $V_\ell$ and 
a subset of unlabeled nodes $V_{u}$. 
The labels for $V_{\ell}$ are represented by a label indicator matrix 
$Y^\ell\in\{0,1\}^{N_\ell\times C}$ where $C$ is the number of classes
and $N_\ell$ is the number of labeled nodes. 

\begin{figure}[!t]
\centering
\includegraphics[width =.95 \textwidth]{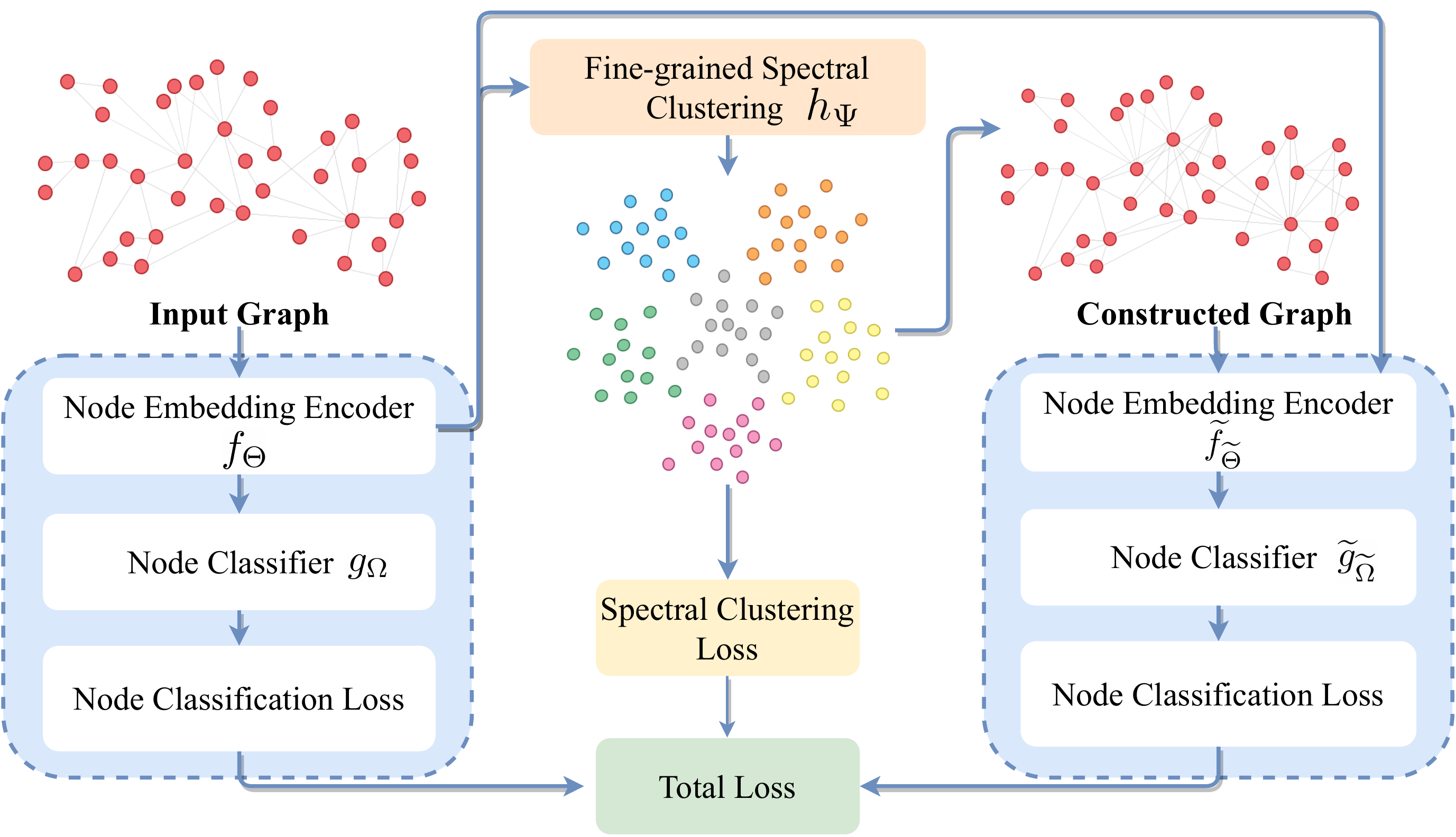} 
\caption{An illustration of the proposed Dual GNN learning framework. The framework contains two modules: The primary module is on the left side
	and the auxiliary module is on the right side. }
\label{fig:diagram}
\end{figure}
\subsection{Dual GNN Learning Framework}

In this section, we present the proposed Dual GNN learning framework for semi-supervised node classification,
which aims to empower a given standard GNN base model to handle the difficult learning scenarios
with scarce labeled nodes and noisy graph structures.
The framework, as shown in Figure \ref{fig:diagram}, is made up of two GNN based node prediction modules,
each of which consists of its own node embedding encoder $f$ and node classifier $g$.
The first module can be treated as a standard {\em primary GNN model}, which 
takes the initial node embedding matrix $X$ and adjacency matrix $A$ as input. 
It produces our primary node classifier which aims to learn discriminative node embeddings by propagating messages from the labeled nodes to the rest of the graph using the input adjacency matrix. 
However, as aforementioned, due to the scarcity of labeled nodes and noisy graph structure, 
the first GNN module is only able to propagate the messages to the local neighbourhood 
of the labeled nodes, while failing to propagate messages to a large portion of the graph. 
The second GNN module is designed to address this issue by constructing a new graph adjacency matrix
with a fine-grained spectral clustering method based on the node embedding results of the first module. 
The new adjacency matrix aims to enable effective message propagation across the whole graph
to induce new node embeddings and classifier. It can be treated as an {\em auxiliary module}.
The two GNN modules are trained in a joint learning framework, 
where the learned node embedding of the first primary
GNN module is updated based on both the local and global information obtained from the primary node classifier and auxiliary node classifier respectively.
Below we present the details of the two modules and the fine-grained clustering component.

\subsubsection{Primary GNN Module for Node Prediction}

This module maintains the learning capacity of a standard GNN baseline with the original graph information.
It takes the initial node embedding $X$ and adjacency matrix $A$ as input 
to learn node embeddings and class prediction probabilities as follows:
\begin{equation}
    H = f_{\Theta}(X,A),\qquad P = g_{\Omega}(H)
\end{equation}
where $f_{\Theta}$ denotes the GNN encoder that takes the adjacency matrix $A$ and the initial node features $X$ as input and uses message passing and message aggregation to learn a new embedding matrix $H\in\mathbb{R}^{N\times d}$ for the nodes. 
$g_{\Omega}$ denotes the node classifier 
which takes the learned node embedding $H$ as input and outputs class prediction probability matrix $P$ of the nodes. 
$\Theta$ and $\Omega$ denote the model parameters for $f$ and $g$ respectively. 
In general, this module can be any GNN model developed in the literature that performs standard graph neural network learning.
It can be trained by minimizing the following supervised node classification loss (cross-entropy loss):
\begin{equation}
	\mathcal{L}_{CE} =  \sum\nolimits_{i \in V_{\ell}} \ell_{CE}(P_i,Y^\ell_i)
\end{equation}
where 
$\ell_{CE}$ denotes the cross-entropy function, $P_i$ and $Y^\ell_i$ denote the predicted class probability vector and 
the true label indicator vector of node $i$ respectively.

\subsubsection{Fine-Grained Spectral Clustering}
To build the auxiliary module, we deploy
a fine-grained spectral clustering component to construct a new adjacency matrix on the input graph. 
The spectral clustering function $h$ is a multi-layer perceptron 
that takes the node embedding matrix $H$ learned from the primary GNN module as input
and returns 
a fine-grained soft-clustering assignment on the graph nodes: 
\begin{equation}
            S = h_{\Psi}(H)
\end{equation}
where $S\in\mathbb{R}^{N\times K}$ is the node soft clustering assignment matrix, $K$ is the number of clusters,
and $\Psi$ denotes the parameters of the spectral clustering function $h$.
As each class could contain multiple clusters, we consider fine-grained clustering 
with a $K$ value that is much larger than the number of classes $C$. 

The fine-grained clustering 
function $h_\Psi$ 
is learned by minimizing the following relaxed min-cut spectral clustering loss function $\mathcal{L}_{sc}$: 
\begin{equation}
    \mathcal{L}_{sc} = - \frac{Tr(S^{T} \tilde{A} S)}{Tr(S^{T} \tilde{D} S)} + \Bigg\| \frac{S^TS}{\|S^TS\|_{F}} - \frac{I_{K}}{\sqrt{K}} \Bigg\|_{F}
\end{equation}
where $\|.\|_F$ denotes the Frobenius norm, 
$\tilde{A}= D^{-\frac{1}{2}}A D^{-\frac{1}{2}}$ is the normalized (original input) adjacency matrix, 
and $D$ and $\tilde{D}$ are the degree matrices of $A$ and $\tilde{A}$ respectively,
which are 
calculated as follows:
\begin{equation}
    D_{ii}  = \sum\nolimits_j A_{ij}, \qquad \tilde{D}_{ii}  = \sum\nolimits_j \tilde{A}_{ij}
\end{equation}
We adopt this relaxed min-cut loss function $\mathcal{L}_{sc}$ from \cite{bianchi2020spectral}
to perform fine-grained clustering learning through $h_\Psi$.
This formulation offers several advantages over the traditional spectral clustering approach: First, it does not require the expensive calculation of the eigen-decomposition of the graph Laplacian.
Second, it takes both the node embedding $H$ and the original graph connectivity structure $A$ into account. 
Minimizing the first term of $\mathcal{L}_{sc}$ pushes 
strongly connected nodes to be clustered together, while enforcing the cluster assignment to 
be learnable from $H$ with function $h_\Psi$. 
The second term of $\mathcal{L}_{sc}$ is a regularization term that penalizes degenerate clustering assignments, and pushes $h_{\Psi}$ to generate non-overlapping orthogonal clusters of relatively similar sizes.

\subsubsection{Auxiliary GNN Module for Node Prediction}

To deploy an auxiliary dual module, we first use 
the learned clustering assignment matrix $S$ to construct a new adjacency matrix $A_{sc}$ as follows:
\begin{equation}
    A_{sc}(i,j)= \begin{cases}
    r(S_{i,:},S_{j,:}),& \text{if } \, r(S_{i,:},S_{j,:})\geq \alpha \\
    0,              & \text{otherwise}
\end{cases}
\end{equation}
where $r(.,.)$ is the Pearson correlation function that measures
the similarity between the clustering assignment vectors on a pair of nodes, 
and $\alpha$ is a hyper-parameter that 
controls the sparsity of the constructed adjacency matrix.
We expect the fine-grained spectral clustering function 
can capture the global geometric information of the original graph structure $A$ and the learned node embeddings $H$. 
By constructing the adjacency matrix $A_{sc}$ from the clustering assignment matrix $S$,
we aim to address the problems of scarce labeled nodes and 
deleted edges in the original graph structure. 
Specifically, 
as a pair of nodes that are either connected in $A$ or have similar embeddings in $H$
tend to have similar clustering assignments through the spectral clustering function,
it enables $A_{sc}$ not only to maintain the connectivity of the original $A$, but also 
to add more edges based on the embedding similarities of the nodes. 
This helps propagate the local messages and node label information to a larger portion of the graph, 
addressing the underlying local overfitting problem caused by scare labeled nodes and 
corrupted graph structures with many deleted edges.

Given the new adjacency matrix $A_{sc}$, 
the auxiliary 
GNN node prediction module 
has a standard architecture with a GNN encoder $\widetilde{f}$ and a classifier $\widetilde{g}$. 
The encoder $\widetilde{f}$ takes the node embedding $H$ from the primary module 
and the new constructed adjacency matrix $A_{sc}$ as input and outputs a node embedding matrix $\widetilde{H}$,
while the node classifier $\widetilde{f}$ further predicts the classification probability matrix $\widetilde{P}$ on all the nodes:
\begin{equation}
	\widetilde{H} = \widetilde{f}_{\widetilde{\Theta}}(H,A_{sc}),
	\qquad \widetilde{P} = \widetilde{g}_{\widetilde{\Omega}}(\widetilde{H})
\end{equation}
where 
$\widetilde{\Theta}$ and $\widetilde{\Omega}$ denote the model parameters for the encoder and classifier respectively.
Specifically we adopt the encoder of the standard Graph Convolutional Networks (GCN) 
\cite{kipf2016semi} as our encoder $\widetilde{f}$. 
We choose to use GCN encoders due to two reasons:
First, GCNs are simple and efficient with very few learnable parameters. 
Second, unlike other GNN models such as GATs, GCNs allow us to utilize the weights of the constructed adjacency matrix $A_{sc}$.
This auxiliary GNN module can be trained in a standard way by minimizing the node classification loss (cross-entropy loss):

\begin{equation}
	\mathcal{\widetilde{L}}_{CE} =  \sum\nolimits_{i \in V_{\ell}} \ell_{CE}(\widetilde{P}_{i},Y^\ell_i)
\end{equation}
\subsubsection{Joint Dual GNN Learning}
Finally we can integrate the classification loss functions, $\mathcal{L}_{CE}$ and $\mathcal{\widetilde{L}}_{CE}$, 
from the two modules and the clustering loss, $\mathcal{L}_{sc}$, together 
to form a joint dual GNN learning problem:
\begin{equation}
	\label{eq:loss}
	\min_{\Theta,\Omega, \widetilde{\Theta}, \widetilde{\Omega},\Psi} \quad 
	\mathcal{L}   = \mathcal{L}_{CE}  +  \mathcal{\widetilde{L}}_{CE}  +  \mathcal{L}_{sc}   
\end{equation}
Note as the embedding matrix $H$ from the primary module 
is used as inputs for the spectral clustering and the auxiliary module,
hence the auxiliary loss $\mathcal{\widetilde{L}}_{CE}$ and the clustering loss $\mathcal{L}_{sc}$,
are also functions of the primary encoding parameters $\Theta$. 
Within such a joint learning framework, 
the two modules can interactively impact each other through the shared encoder $f_\Theta$.
The entire network is trained end-to-end to minimize $\mathcal{L}$ and the process is presented in Algorithm \ref{alg:method}.

\begin{algorithm}[t!]
    \caption{Training Algorithm for the Dual GNN Learning Framework}
    \label{alg:method}
    \begin{algorithmic}
\STATE{\textbf{Input:} Given graph $G$ with input feature matrix $X$, \\ adjacency matrix $A$, labeled node set $V_{\ell}$; hyper-parameters $\alpha$, $K$}
	    \STATE{\textbf{Output:} Learned model parameters 
	    $\Theta, \Omega$, $\widetilde{\Theta}, \widetilde{\Omega}$, $\Psi$ }
	    \FOR{{iter = 1} {\bf to} maxiters}
        \STATE $H = f_{\Theta}(X,A)$ ,\, compute node embeddings $H$
        \STATE $P = g_{\Omega}(H)$ ,\, compute nodes classification probabilities
        \STATE $S = h_{\Psi}(H)$ ,\, compute the fine-grained spectral clustering of nodes
        \STATE Construct adjacency matrix $A_{sc}$ using $S$
	    \STATE $\widetilde{H} = \widetilde{f}_{\widetilde{\Theta}}(H,A_{sc})$ ,\, compute node embeddings $\widetilde{H}$
	    \STATE $\widetilde{P} = \widetilde{g}_{\widetilde{\Omega}}(\widetilde{H})$ ,\, compute nodes classification probabilities
	    \STATE Calculate $ \mathcal{L}   = \mathcal{L}_{CE}  +  \mathcal{\widetilde{L}}_{CE}  +  \mathcal{L}_{sc}   $
	    \STATE{Update model parameters $\Theta, \Omega$,  $\widetilde{\Theta}, \widetilde{\Omega}$, $\Psi$ 
	    to minimize $\mathcal{L}$ with gradient descent}
    \ENDFOR
    \end{algorithmic}
\end{algorithm}

\section{Experiments}
We tested the proposed Dual GNN learning framework by applying it on multiple GNN baseline models,
and conducted experiments on three different experimental setups: 
learning with few labeled nodes, learning with noisy graph structures, and learning with both few labeld nodes
and noisy graph structures.  In this section, we report the experimental settings and results. 

\subsection{Experiment Settings}
\subsubsection{Datasets \& Baselines}

We used the citation network datasets (Cora, CiteSeer) \cite{sen2008collective} where the nodes in the graph correspond to documents and the edges correspond to citations between documents. The features of each node in the graph are a bag-of-words of the node document, and the label of each node is the subject of the corresponding document. We utilized the same train/validation/test/unlabeled node split provided by \cite{yang2016revisiting}, where the training set is made up of 20 nodes per class, the validation set is made up of 500 nodes, the test set is made up of 1000 nodes, while the remaining nodes in the graph constitute the unlabeled nodes. 
We applied our proposed Dual GNN framework on four GNN baselines
by using each of them as our primary module: Graph Convolution Networks (GCNs) \cite{kipf2016semi}, Graph Attention Networks (GATs) \cite{velivckovic2017graph}, Topology Adaptive Graph Convolutional Networks (TAG) \cite{du2017topology}, and Dynamic Neighborhood Aggregation in Graph Neural Networks (DNA) \cite{fey2019just}.

\subsubsection{Implementation Details}

For the Dual GNN framework and the baselines,
the node embedding encoders are made up of two message-passing layers, 
while the node classifiers contain a single fully connected layer. 
The spectral clustering function $h$ is a one-layer perceptron. 
The non-linear activation function applied on each message-passing layer is 
the exponential linear units (ELUs).
We apply $L_2$ regularization (weight decay = $1e^{-3}$) to all model parameters. 
GAT has a single attention head with an attention dropout rate of $0.5$. 
DNA has an additional single fully-connected layer prior to the message-passing layers to project the initial node embeddings to a lower dimensional space of size $32$. 
All networks are trained for $500$ training epochs with Adam optimizer with a learning rate of $1e^{-2}$ 
and a scheduler (step size $=50$ training iterations) with a learning rate decay factor of $\gamma = 0.5$. 
For the Dual GNN framework, 
the number of clusters is set to $K = 10 \times C$, where $C$ is the number of classes, 
and the sparsity threshold for constructing the new adjacency matrix is $\alpha = 0.7$.
We used the default model parameter initialization of the Pytorch-Geometric  \cite{fey2019fast}.

\subsection{Experiments with Few Labeled Nodes}
In this set of experiments, 
we aim to investigate the performance of the Dual GNN framework 
with significantly fewer labeled nodes. 
The public splits of the datasets contain 20 labeled training nodes per class. 
We tested a number of much smaller label rates by using 
$\{2, 3, 5, 10\}$ labeled nodes per class respectively.
For each label rate, we generate $10$ random subsets of labeled nodes 
from the labeled training set. 
For each random subset, we repeat 5 runs for each comparison method. 
We report the mean test classification accuracy results
and the corresponding standard deviations over the total 50 runs. 
We also conducted experiments by using all the 20 labeled nodes per class.
In this case, the average results over 5 runs are reported. 

We applied the Dual GNN framework over four baseline models. 
The comparison results over each baseline model and the corresponding Dual method are reported in Table \ref{tab:exp_1}. 
We can see that 
the performance of all four baselines (GCN, GAT, TAG, DNA) degrades as the number of labeled nodes per class decreases. 
The performance of DNA baseline degrades more significantly compared to the other three baselines, 
which can be explained by its relatively large number of learnable parameters. 
The TAG baseline performs better than other baselines at low label rates, 
which can be attributed to its inherently large receptive field that 
allows messages to be propagated to a larger portion of the graph \cite{luan2019break}. 
This can also explain the relatively similar performance between the Dual TAG and the TAG baseline.
The proposed framework however consistently improves the performance of all the other three base models, GCN, GAT and DNA,
across all label rates on both datasets.
The performance gains achieved by the Dual framework are particularly remarkable with smaller label rates. 
For example, Dual GCN outperforms GCN by 7.6\% and 8.4\% on Cora and CiteSeer respectively with 2 labeled nodes per class.
This suggests the proposed Dual framework is very beneficial for GNN learning with smaller number of labeled nodes.

\begin{table}[t]
\caption{Mean classification accuracy (standard deviation is within brackets) on Cora (left part) and CiteSeer (right part) datasets with few labeled nodes.}
\vskip .1in
\renewcommand{\arraystretch}{1.1}
    \resizebox{\textwidth}{!}{
\begin{tabular}{l|c c c c c||c c c c c}
\hline
& \multicolumn{5}{c||}{\bf Cora} & \multicolumn{5}{c}{\bf CiteSeer}\\
& 2 & 3 & 5 & 10 & 20 & 2 & 3 & 5 & 10 & 20 \\
\hline
	{ GCN}   & $63.5_{(4.4)}$ & $69.2_{(3.5)}$  & $75.1_{(3.1)}$    & $78.8_{(0.9)}$   & $81.5_{(0.6)}$ & $48.1_{(7.9)}$ & $56.4_{(6.2)}$    & $62.2_{(3.5)}$   & $66.3_{(1.8)}$    & $68.5_{(0.7)}$       \\

	{ Dual GCN} & $\textbf{71.1}_{(4.0)}$  & $\textbf{74.8}_{(2.4)}$ &  $\textbf{77.6}_{(2.6)}$    & $\textbf{79.9}_{(1.0)}$    & $\textbf{82.7}_{(0.5)}$    & $\textbf{56.5}_{(9.1)}$ & $\textbf{63.0}_{(5.7)}$  & $\textbf{65.4}_{(4.1)}$     & $\textbf{66.5}_{(2.4)}$  & $\textbf{70.0}_{(1.5)}$              \\
\hline
\hline
	{ GAT}      & $63.5_{(3.9)}$   & $69.6_{(3.4)}$   & $74.0_{(3.5)}$  & $79.0_{(1.2)}$     & $81.3_{(0.5)}$  & $49.2_{(8.1)}$     & $56.0_{(6.6)}$      & $61.8_{(3.3)}$             & $65.2_{(1.9)}$    & $69.0_{(0.6)}$              \\

	{ Dual GAT} & $\textbf{65.7}_{(5.2)}$  & $\textbf{71.6}_{(2.6)}$    & $\textbf{75.4}_{(2.7)}$    & $\textbf{79.1}_{(1.4)}$    & $\textbf{81.6}_{(0.3)}$    & $\textbf{50.8}_{(7.3)}$     & $\textbf{58.5}_{(5.3)}$     & $\textbf{63.2}_{(3.5)}$     & $\textbf{65.9}_{(2.0)}$   & $\textbf{69.0}_{(0.5)}$              \\
\hline
\hline
	{ TAG}      & $\textbf{67.7}_{(4.3)}$   & $73.0_{(2.7)} $    & $76.3_{(2.8)}$    & $79.9_{(1.2)}$  & $\textbf{82.6}_{(0.4)}$     & $\textbf{50.7}_{(9.8)}$    & $\textbf{58.5}_{(6.6)}$ & $\textbf{63.6}_{(3.1)}$  & $\textbf{67.7}_{(1.3)}$  & $\textbf{70.0}_{(0.9)}$              \\

	{ Dual TAG} & $67.5_{(6.8)}$    & $\textbf{75.2}_{(3.5)}$ & $\textbf{77.5}_{(2.6)}$    &  $\textbf{80.0}_{(1.5)}$   & $82.0_{(0.5)}$             & $49.0_{(1.1)}$ & $57.2_{(8.5)}$     & $61.9_{(4.3)}$     & $64.7_{(2.6)}$      & $67.9_{(1.2)}$   \\
\hline
\hline
	{ DNA}    & $56.7_{(5.3)}$   & $65.1_{(3.2)}$     & $70.3_{(3.3)}$     & $77.0_{(1.2)}$  & $\textbf{81.3}_{(0.5)}$    & $43.9_{(7.1)}$    & $53.0_{(5.8)}$ & $58.6_{(3.5)}$     & $64.4_{(2.1)}$     & $69.5_{(0.7)}$              \\

	{ Dual DNA} & $\textbf{62.8}_{(5.0)}$   & $\textbf{69.2}_{(3.5)}$   & $\textbf{74.0}_{(3.0)}$     & $\textbf{77.6}_{(1.2)}$  & $81.0_{(0.3)}$ & $\textbf{51.5}_{(8.0)}$             & $\textbf{60.8}_{(5.0)}$             & $\textbf{64.3}_{(3.5)}$             & $\textbf{67.1}_{(2.1)}$              & $\textbf{70.1}_{(0.6)}$ \\           
\hline
\end{tabular}}
    \label{tab:exp_1}

\end{table}

\begin{figure}[!t]
\centering
\includegraphics[width=.6\textwidth]{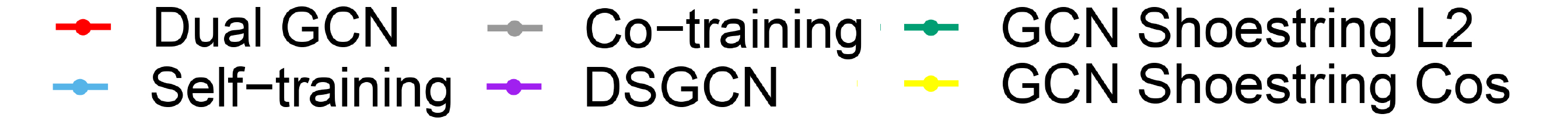}
\begin{subfigure}{0.495\textwidth}
\includegraphics[width = .8\textwidth, height=1.6in]{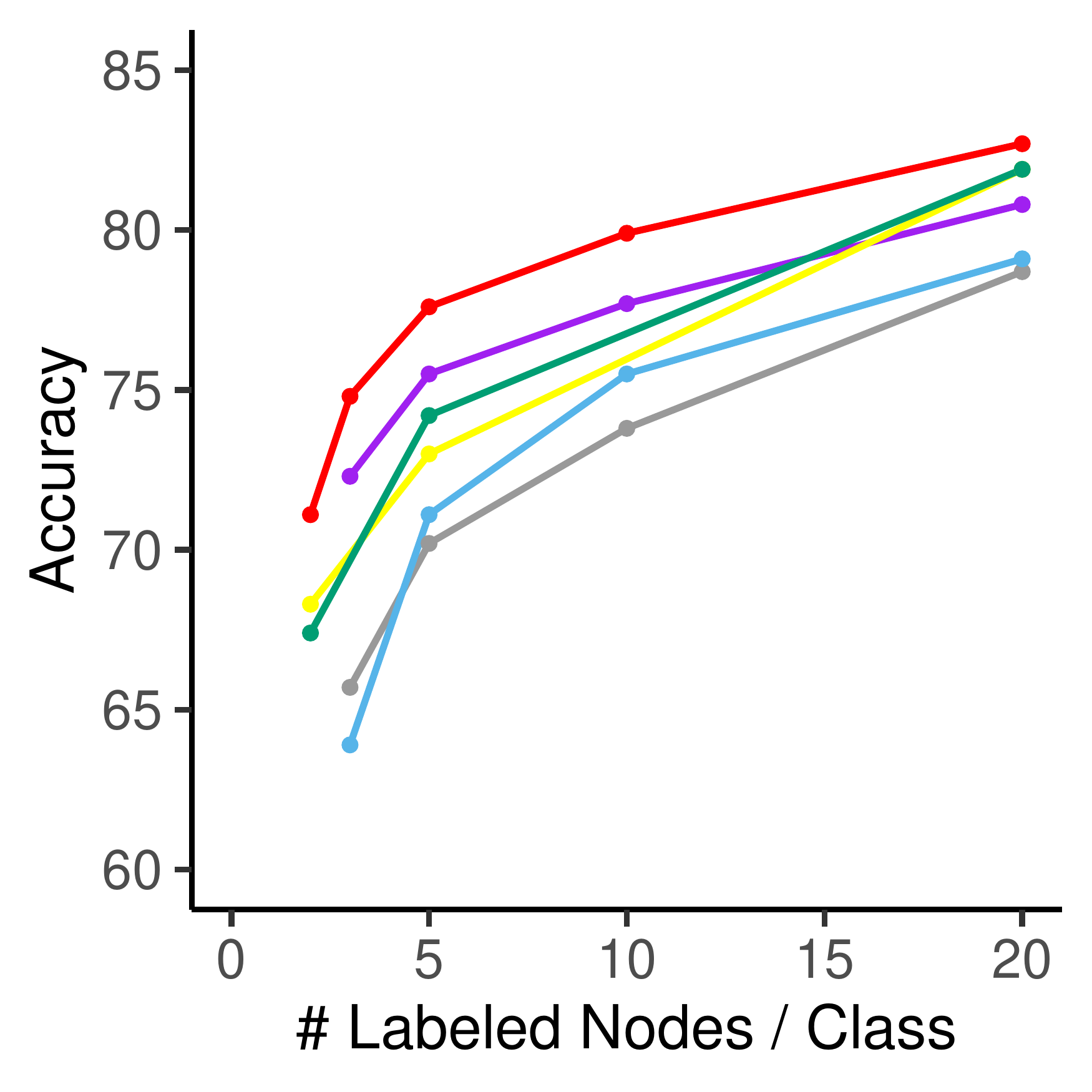}
\caption{Cora}
\end{subfigure}
\begin{subfigure}{0.495\textwidth}
\includegraphics[width = .8\textwidth, height=1.6in]{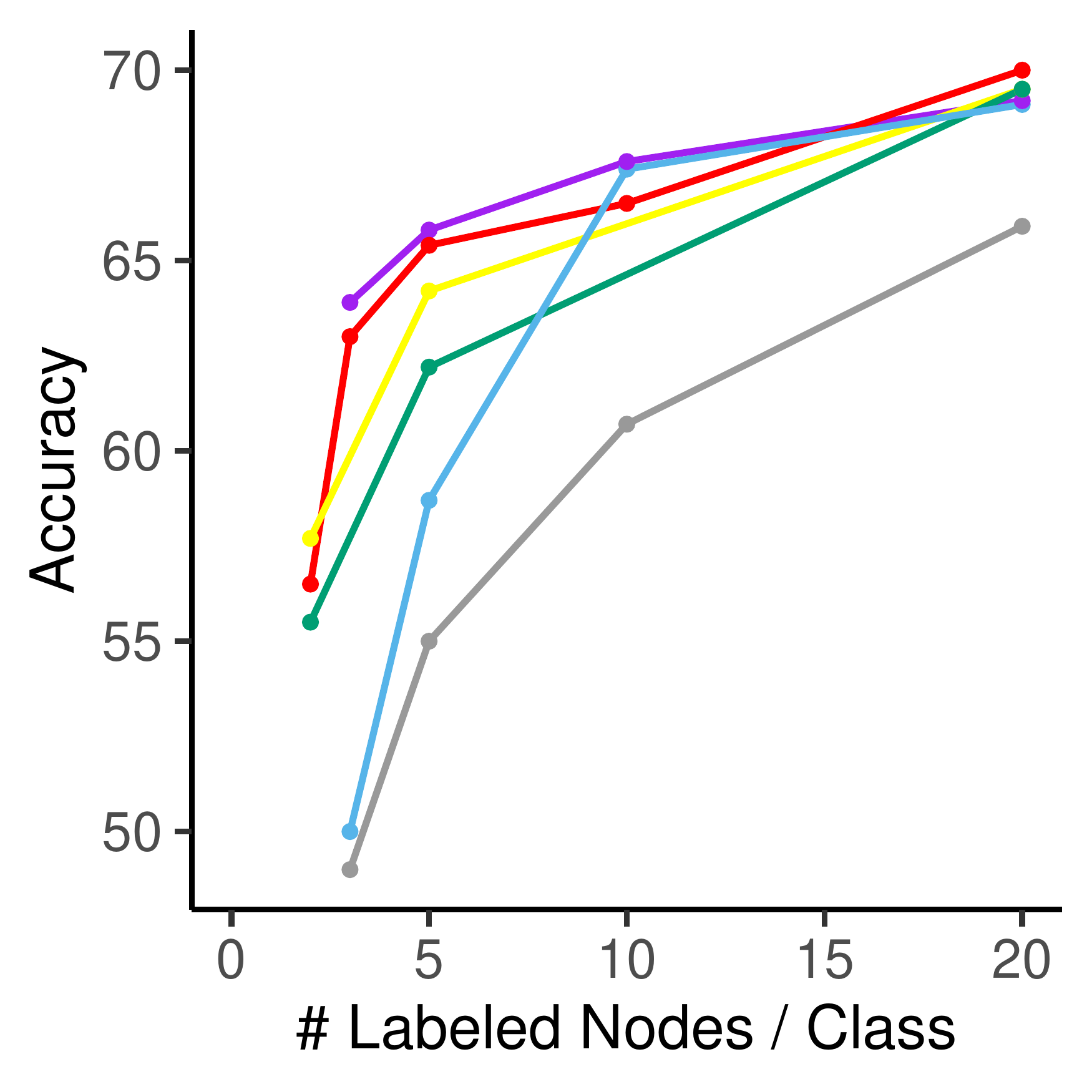}
\caption{CiteSeer}
\end{subfigure}
\caption{Mean classification accuracy on Cora (left part) and CiteSeer (right part) datasets with few labeled nodes.}
\label{fig:compare}
\end{figure}

We also compared the proposed Dual GNN framework 
with a number of methods developed in the literature that address the degradation of performance of GNNs 
with few labeled nodes, including two Shoestring methods (GCN Shoestring L2, GCN Shoestring Cos) \cite{lin2020shoestring} and three Dynamic self-training methods (Co-training, Self-training, DSGCN) \cite{zhou2019dynamic}. 
We used the reported results from \cite{lin2020shoestring} and \cite{zhou2019dynamic}. 
We used GCN as the primary module and compared our proposed Dual GCN with these five comparison methods
	with different label rates. The comparison results on the two datasets, Cora and CiteSeer,
	are reported in Figure \ref{fig:compare}.
We can see that 	
Dual GCN consistently outperforms all the other methods across all label rates 
with notable performance gains on the Cora dataset -- 
the average performance gain is about $+2\%$. 
On the CiteSeer dataset, although Dual GCN is outperformed by DSGCN, the performance difference
between them is very small, while Dual GCN outperforms the other comparison methods.

\begin{table}[t]
\caption{Mean classification accuracy (standard deviation is within brackets) on Cora (left part) and CiteSeer (right part) datasets with 
	    different edge corruption rates, (0.25, 0.5, 0.75, 0.9, 0.95). 
	    }
\vskip .1in	    
\renewcommand{\arraystretch}{1.1}
     \resizebox{\textwidth}{!}{
\begin{tabular}{l|c c c c c||c c c c c}
\hline
& \multicolumn{5}{c||}{\bf Cora} & \multicolumn{5}{c}{\bf CiteSeer}\\
	&  0.25 & 0.5 & 0.75 & 0.90 & 0.95 & 0.25 & 0.5 & 0.75 & 0.90 & 0.95 \\
\hline
	{ GCN}                         & $79.1_{(1.0)}$                & $74.7_{(1.5)}$               & $67.4_{(1.8)}$                & $60.1_{(1.7)}$              & $57.9_{(1.6)}$                 & $67.2_{(1.1)}$                & $64.5_{(1.1)}$               & $58.7_{(1.7)}$                & $55.7_{1.8}$               & $54.9_{(1.7)}$                \\

	{ Dual GCN}                  & $\textbf{79.9}_{(0.8)}$                 & $\textbf{75.9}_{(1.5)}$               & $\textbf{69.0}_{(1.7)}$                & $\textbf{64.1}_{(2.0)}$                & $\textbf{60.8}_{(1.8)}$                & $\textbf{67.7}_{(1.4)}$                & $\textbf{66.2}_{(2.6)}$               & $\textbf{64.2}_{(3.1)}$                 & $\textbf{60.6}_{(3.1)}$                & $\textbf{61.4}_{(2.1)}$                 \\
\hline
\hline
	{ GAT}                          & $78.7_{(1.1)}$                & $75.2_{(1.3)}$              & $68.9_{(1.8)}$                & $\textbf{61.4}_{(1.8)}$               & $58.2_{(1.6)}$               & $67.2_{(1.1)}$                & $65.1_{(1.2)}$               & $59.4_{(1.9)}$                & $56.4_{(2.3)}$               & $54.3_{(2.3)}$                \\

	{ Dual GAT}                 & $\textbf{78.9}_{(0.8)}$                & $\textbf{75.3}_{(1.5)}$               & $\textbf{68.9}_{(1.6)}$                & $60.6_{(2.3)}$               & $\textbf{59.3}_{(2.3)}$               & $\textbf{67.6}_{(1.3)}$                & $\textbf{65.2}_{(1.6)}$               & $\textbf{59.6}_{(1.9)}$                & $\textbf{57.3}_{(2.7)}$               & $\textbf{59.2}_{(2.2)}$                \\
\hline
\hline
	{ TAG}             & $77.6_{(1.0)}$                & $70.6_{(1.6)}$               & $60.0_{(1.9)}$               & $55.3_{(1.6)}$               & $54.6_{(1.6)}$                & $\textbf{66.3}_{(1.1)}$                & $62.2_{(1.1)}$               & $55.3_{(0.9)}$                & $53.9_{(1.2)}$               & $53.9_{(0.8)}$                \\

	{ Dual TAG}                & $\textbf{78.8}_{(1.5)}$                & $\textbf{72.4}_{(1.7)}$               & $\textbf{64.2}_{(2.0)}$               & $\textbf{57.4}_{(1.6)}$               & $\textbf{57.8}_{(2.1)}$                 & $65.0_{(1.8)}$                & $\textbf{62.6}_{(2.8)}$               & $\textbf{57.7}_{(3.3)}$                & $\textbf{57.8}_{(4.3)}$               & $\textbf{60.6}_{(2.5)}$                \\
\hline
\hline
	{ DNA}                     & $78.7_{(0.7)}$                & $74.9_{(1.1)}$               & $68.9_{(1.5)}$                & $62.6_{(1.8)}$               & $60.3_{(1.0)}$               & $67.6_{(1.2)}$                & $65.5_{(1.1)}$               & $60.7_{(1.1)}$                & $57.9_{(1.3)}$               & $57.1_{(1.3)}$                \\

	{ Dual DNA}             & $\textbf{79.0}_{(1.3)}$                & $\textbf{76.0}_{(1.3)}$                & $\textbf{70.2}_{(1.3)}$                 & $\textbf{63.9}_{(1.9)}$               & $\textbf{61.6}_{(1.5)}$                & $\textbf{68.9}_{(1.2)}$                 & $\textbf{67.6}_{(1.1)}$                & $\textbf{62.9}_{(1.8)}$                & $\textbf{60.0}_{(2.2)}$               & $\textbf{60.8}_{(1.9)}$                \\
\hline
\end{tabular}}
    \label{tab:exp_2}

\end{table}

\subsection{Experiments with Noisy Graph Structures}

In the second set of experiments, we investigate the robustness of the proposed Dual GNN framework
to noisy/corrupt graph structures with missing edges.
We simulate untargeted adversarial attacks by randomly deleting edges from the input graphs, 
which is similar to the adversarial setups in \cite{elinas2019variational,geisler2020reliable,chen2020iterative}.
Specifically, to corrupt the input graph structure, we randomly drop a portion of the edges in the adjacency matrix of the graph.
We considered a set of different edge drop ratios, i.e., corruption rates: $\{0.25, 0.50, 0.75, 0.90, 0.95\}$. 
For each corruption rate, we generate $10$ corrupted input adjacency matrices and train on each one of the resulting graphs $5$ times. We report the mean classification accuracy and the corresponding standard deviation 
across all runs for each corruption rate. 

Again we applied the proposed framework on four baseline models. 
The comparison results on the two datasets are reported in Table \ref{tab:exp_2}.
The table shows the vulnerability of GNNs to attacks on the underlying graph structures 
where the performance of all GNN baselines declines substantially as 
the edge corruption ratio increases. 
This is due to the dependence of GNNs on the graph structures for propagating messages/labels across the graph.
The TAG baseline is particularly sensitive to noise in graph structures due to its large but static receptive field, 
which is clearly demonstrated by its notably larger performance drop relative to the other baselines. 
Nevertheless, the proposed Dual learning framework consistently improves the performance of 
all four 
GNN baselines across all corruption rates on both benchmark datasets. 
The performance gain of our framework grows as the percentage of deleted edges increases. 
These results demonstrate the wide applicability of the proposed framework in alleviating 
the negative impact of edge corruptions.

\begin{table}[t]
\caption{Mean classification accuracy (standard deviation is within brackets) on Cora (left part) and CiteSeer (right part) datasets with few labeled nodes \& noisy graph structures.}
\vskip .1in
\renewcommand{\arraystretch}{1.1}
\resizebox{\textwidth}{!}{%
\begin{tabular}{l|l|c c c c c||c c c c c }
\hline
            &        & \multicolumn{5}{c||}{\bf Cora}& \multicolumn{5}{c}{\bf CiteSeer}            \\
& & 0.25                  & 0.5                   & 0.75                  & 0.9                   & 0.95                  & 0.25                  & 0.5                   & 0.75                   & 0.9                   & 0.95                  \\
\hline
\multirow{4}{*}{\rotatebox[origin=c]{90}{3 Labels} }&GCN                 & $65.6_{(3.5)}$          & $58.2_{(4.3)}$          & $48.4_{(3.6)}$          & $40.2_{(3.8)}$          & $39.4_{(3.9)}$          & $54.4_{(5.4)}$          & $50.5_{(5.7)}$          & $42.1_{(4.3)}$           & $40.9_{(2.9)}$          & $39.9_{(3.0)}$          \\

& Dual GCN            & $\textbf{72.4}_{(2.3)}$ & $\textbf{64.7}_{(4.5)}$ & $\textbf{54.3}_{(4.8)}$ & $\textbf{41.9}_{(6.8)}$ & $\textbf{40.0}_{(7.3)}$ & $\textbf{62.2}_{(4.9)}$ & $\textbf{55.0}_{(8.8)}$ & $\textbf{46.7}_{(11.0)}$ & $\textbf{41.8}_{(8.9)}$ & $\textbf{42.7}_{(6.9)}$ \\
& DNA                 & $61.5_{(3.8)}$          & $54.9_{(4.6)}$          & $47.6_{(4.2)}$          & $41.7_{(4.0)}$          & $39.8_{(4.2)}$          & $52.6_{(4.6)}$          & $49.6_{(4.7)}$          & $43.4_{(4.5)}$           & $41.8_{(2.8)}$          & $40.8_{(3.8)}$          \\
& Dual DNA            & $\textbf{66.3}_{(3.7)}$ & $\textbf{60.3}_{(4.4)}$ & $\textbf{52.0}_{(5.1)}$ & $\textbf{43.2}_{(4.6)}$ & $\textbf{41.3}_{(4.1)}$ & $\textbf{59.1}_{(4.8)}$ &  $\textbf{52.2}_{(6.1)}$    &         $\textbf{45.6}_{(5.5)}$  & $\textbf{42.7}_{(4.3)}$ & $\textbf{42.2}_{(4.0)}$ \\
\hline \hline
\multirow{4}{*}{\rotatebox[origin=c]{90}{5 Labels} }& GCN                 & $71.4_{(3.2)}$          & $64.5_{(5.5)}$          & $56.1_{(3.3)}$          & $45.0_{(3.3)}$          & $42.0_{(3.0)}$          & $59.0_{(3.6)}$          & $53.6_{(4.5)}$          & $47.6_{(3.8)}$           & $44.3_{(3.8)}$          & $44.7_{(2.6)}$          \\
& Dual GCN            & $\textbf{74.9}_{(3.0)}$ & $\textbf{68.8}_{(4.8)}$ & $\textbf{60.9}_{(3.2)}$ & $\textbf{47.8}_{(7.0)}$ & $\textbf{44.1}_{(6.9)}$ & $\textbf{62.4}_{(5.3)}$ & $\textbf{58.8}_{(6.6)}$ & $\textbf{53.1}_{(1.0)}$  & $\textbf{45.7}_{(9.7)}$ & $\textbf{46.4}_{(9.7)}$ \\
& DNA                 & $67.2_{(3.3)}$          & $61.4_{(5.1)}$          & $54.1_{(2.7)}$          & $46.2_{(2.6)}$          & $44.2_{(2.3)}$          & $57.1_{(2.9)}$          & $53.2_{(3.8)}$          & $48.2_{(3.4)}$           & $46.2_{(3.1)}$          & $45.2_{(2.6)}$          \\
& Dual DNA       & $\textbf{70.2}_{(3.6)}$   & $\textbf{65.8}_{(4.9)}$   & $\textbf{58.3}_{(3.2)}$  & $\textbf{48.4}_{(4.6)}$   & $\textbf{44.9}_{(3.7)}$ & $\textbf{62.0}_{(4.0)}$  & $\textbf{57.5}_{(4.9)}$   & $\textbf{50.9}_{(4.8)}$  &  $\textbf{47.6}_{(3.8)}$ & $\textbf{47.7}_{(3.8)}$  \\
\hline \hline
\multirow{4}{*}{\rotatebox[origin=c]{90}{10 Labels} }& GCN    & $76.2_{(0.9)}$ & $69.9_{(1.7)}$   & $62.1_{(2.5)}$        & $54.1_{(2.7)}$      & $52.7_{(2.6)}$      & $63.3_{(2.4)}$     & $59.4_{(3.6)}$  & $52.9_{(3.3)}$ & $50.1_{(3.5)}$     & $50.4_{(2.5)}$   \\
& Dual GCN    & $\textbf{77.7}_{(1.1)}$ & $\textbf{72.8}_{(1.7)}$  & $\textbf{65.4}_{(2.7)}$ & $\textbf{57.9}_{(4.0)}$  & $\textbf{56.6}_{(4.0)}$  & $\textbf{64.8}_{(2.9)}$   &  $\textbf{62.9}_{(3.7)}$  & $\textbf{59.5}_{(5.5)}$   & $\textbf{54.0}_{(6.5)}$  & $\textbf{55.8}_{(5.2)}$ \\
& DNA   & $74.3_{(1.5)}$   & $69.2_{(1.2)}$    & $63.3_{(1.8)}$   & $56.5_{(1.9)}$  & $55.2_{(2.3)}$      & $62.6_{(2.5)}$         & $59.7_{(2.9)}$  & $54.1_{(3.2)}$ & $52.3_{(3.0)}$  & $52.1_{(2.3)}$      \\
& Dual DNA       & $\textbf{75.9}_{(1.3)}$  & $\textbf{71.0}_{(1.3)}$   & $\textbf{65.7}_{(1.6)}$  & $\textbf{58.2}_{(2.2)}$  & $\textbf{55.8}_{(2.4)}$ & $\textbf{66.0}_{(2.3)}$ & $\textbf{63.7}_{(2.5)}$ & $\textbf{56.8}_{(3.0)}$   &  $\textbf{53.6}_{(3.3)}$ & $\textbf{55.1}_{(3.0)}$  \\
\hline
\end{tabular}}
    \label{tab:exp_3}
\end{table}

\subsection{Experiments with Few Labeld Nodes \& Noisy Graph Structures}

We also conducted further experiments to  
investigate the performance of our proposed Dual learning framework with both few labeled nodes and noisy graph structures. 
We used three label rates with $\{3,5,10\}$ labeled nodes per class respectively.
For each label rate, we create $10$ different training sets as before. 
We test the performance of our proposed framework with two baselines, GCN and DNA, 
	under different graph corruption rates from $\{0.25, 0.50, 0.75, 0.90, 0.95\}$.
We report the mean classification accuracy and standard deviation results for each rate across all corruption rates.

Table \ref{tab:exp_3} presents the obtained comparison results, 
It includes three sections of results.
The top section of the table reports the comparison results with different edge drop ratios for 3 labeled nodes per class, 
the middle section reports the comparison results for 5 labeled nodes per class,
and the bottom section 
reports the results for 10 labeled nodes per class. 
We can see that with combination of small label rates and large edge corruption ratios causes 
substantial performance drops on the baseline models. 
Our proposed framework consistently improves the performance of both GCN and DNA across both datasets, 
all label rates and all graph corruption ratios. 
This again validated the general efficacy of the proposed Dual GNN framework.

\begin{figure*}[h]
\centering
\includegraphics[width = 0.60 \textwidth]{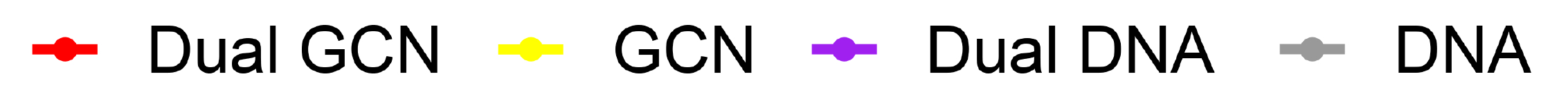}\\
\begin{subfigure}{0.48\textwidth}
\includegraphics[width=0.8\textwidth,height=1.6in]{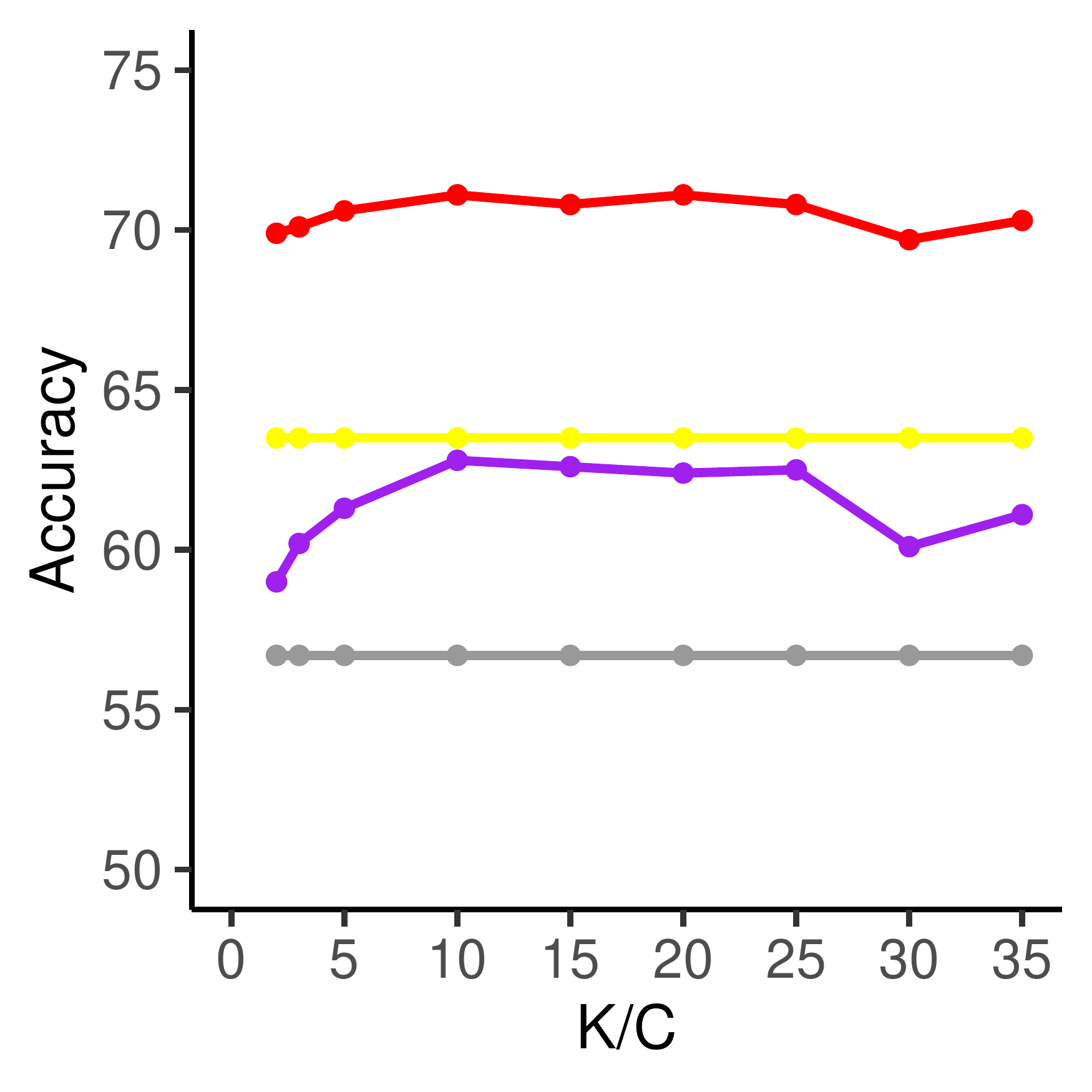} 
\caption{Cora, K}
\label{fig:sen_a}
\end{subfigure}
\begin{subfigure}{0.48\textwidth}
\includegraphics[width = 0.8\textwidth,height=1.6in]{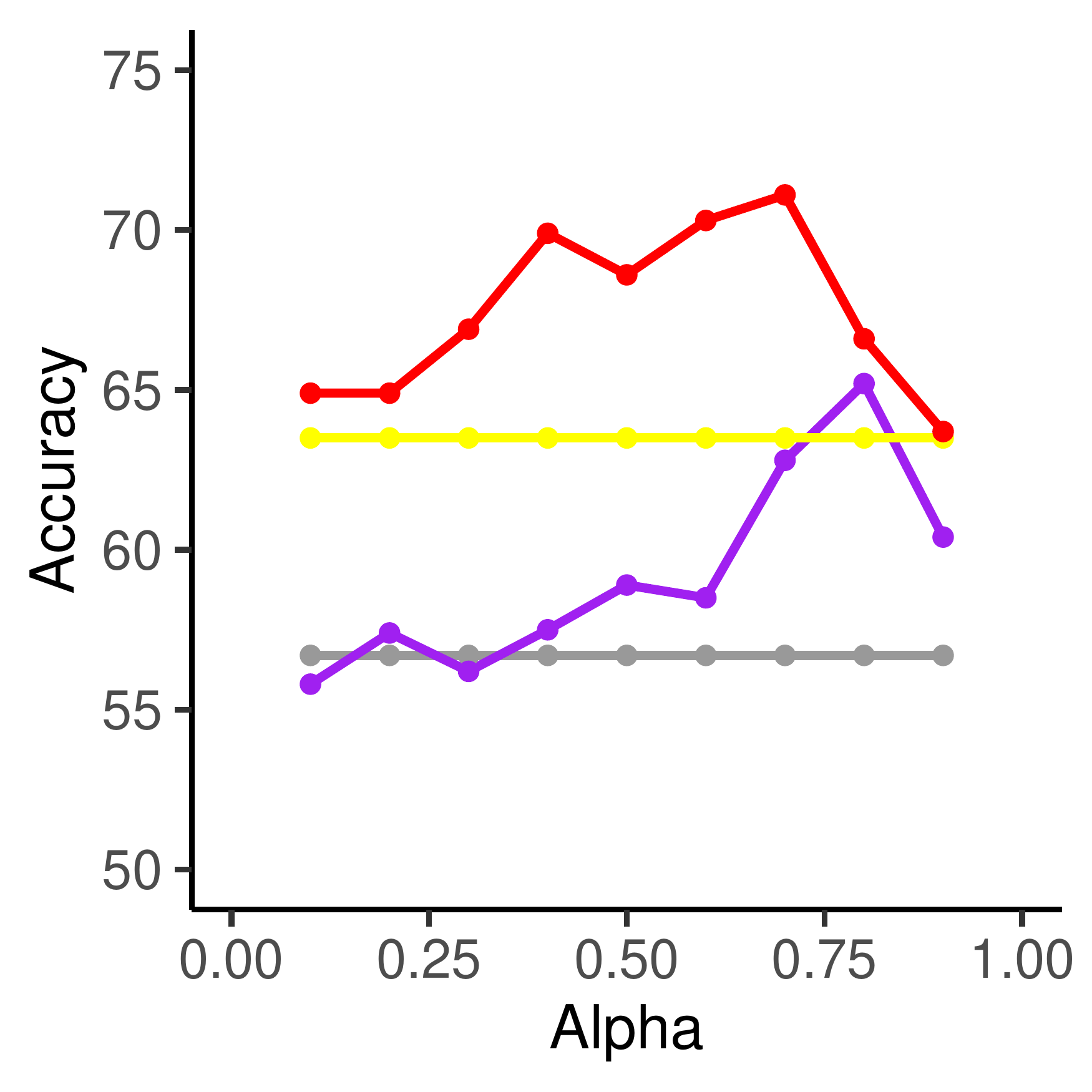} 
\caption{Cora, $\alpha$}
\label{fig:sen_c}
\end{subfigure}
\caption{Sensitivity analysis for the proposed Dual framework on hyper-parameters $K$ and $\alpha$: 
	(a) Cora, $K$, (b) Cora, $\alpha$.}
\label{fig:Sensitivity}
\end{figure*}

\subsection{Hyper-parameter Sensitivity Analysis}

We also investigated the influence of the two hyper-parameters of the proposed framework on its performance: 
the number of clusters $K$ and the sparsity threshold $\alpha$.
We consider a set of $K$ values from $K/C=\{2,3,5,10,15,20,25,30,35\}$
and $\alpha$ values from $\{0.1,0.2,0.3,0.4,0.5,0.6,0.7,0.8,0.9\}$.
Figure \ref{fig:Sensitivity} shows the comparison results of the proposed framework with the GCN and DNA baselines 
on the Cora dataset by varying the values of $K$ or $\alpha$ separately. 

From Figures \ref{fig:sen_a} we can see that the 
proposed framework is resilient to changes in the number of cluster $K$
within a large range of values. 
Both Dual GCN and Dual DNA substantially outperform the underlying baselines across all K/C values. 
Choosing a very small value for K may cause dissimilar nodes to be clustered together which induces a larger number of inter-class edges and can possibly impair the performance, 
while choosing a very large value for K can cause fragmented clusters where similar nodes have dissimilar cluster assignments and hence negatively impair the new adjacency matrix. 
A reasonable choice of values for $K/C$ can be between 10 and 25. 
The best value of $K$ can be dataset dependent and hence 
can be selected using a validation set. 

As for the sparsity threshold $\alpha$, the proposed framework is quite sensitive to changes in the $\alpha$ value. 
As shown in Figures \ref{fig:sen_c}, 
A very small 
$\alpha$ value leads to poor performance.
This is because a small $\alpha$ can induce too many inter-class edges in the constructed adjacency matrix 
which damages the classification performance of the proposed framework. 
However, assigning a very large value to $\alpha$ can induce
a very sparse adjacency matrix, which can prevent the framework from propagating the messages across the entire graph
and hence fail the purpose of the proposed framework. 
It is important to maintain a balance between producing a too sparse or a too dense adjacency matrix.
It is a reasonable to choose a $\alpha$ value between 0.6 and 0.8.

\begin{table}[t]
\caption{Ablation study results 
in terms of mean classification accuracy (standard deviation is within brackets)
	on Cora and CiteSeer with few labeled nodes. 
	``Prim.+C" and ``Aux.+C" are abbreviations for the two variants ``Primary+Cluster" and ``Auxiliary+Cluster" respectively.
	}
\vskip .1in
\renewcommand{\arraystretch}{1.1}
    \resizebox{\textwidth}{!}{
\begin{tabular}{l|c c c c c||c c c c c}
\hline
& \multicolumn{5}{c||}{\bf Cora} & \multicolumn{5}{c}{\bf CiteSeer}\\
 & 2 & 3 & 5 & 10 & 20 & 2 & 3 & 5 & 10 & 20 \\
\hline

Dual GCN & $\textbf{71.1}_{(4.0)}$  & $\textbf{74.8}_{(2.4)}$ &  $\textbf{77.6}_{(2.6)}$    & $\textbf{79.9}_{(1.0)}$    & $\textbf{82.7}_{(0.5)}$    & $\textbf{56.5}_{(9.1)}$ & $\textbf{63.0}_{(5.7)}$  & $\textbf{65.4}_{(4.1)}$     & $\textbf{66.5}_{(2.4)}$  & $\textbf{70.0}_{(1.5)}$              \\
GCN   & $63.5_{(4.4)}$ & $69.2_{(3.5)}$  & $75.1_{(3.1)}$    & $78.8_{(0.9)}$   & $81.5_{(0.6)}$ & $48.1_{(7.9)}$ & $56.4_{(6.2)}$    & $62.2_{(3.5)}$   & $66.3_{(1.8)}$    & $68.5_{(0.7)}$       \\
Prim.+C & $63.1_{(5.0)}$ &
$69.6_{(3.1)}$  &  $75.1_{(3.3)}$   & $78.6_{(1.2)}$  &
$81.6_{(0.4)}$    & $49.0_{(7.2)}$ & $57.2_{(5.7)}$ & $62.7_{(3.4)}$ &
$66.0_{(1.9)}$ & $69.2_{(0.4)}$ \\
Aux.+C &   $25.6_{(9.1)}$
&$24.4_{(9.7)}$  & $25.3_{(11.0)}$    & $22.4_{(9.9)}$    &
$24.6_{(8.9)}$ & $26.0_{(8.4)}$  & $27.1_{(9.1)}$     & $27.6_{(8.4)}$
& $28.7_{(10.1)}$ & $25.6_{(7.4)}$   \\
\hline
%
\end{tabular}}
    \label{tab:ablation}
\end{table}

\subsection{Ablation Study}
We further conducted 
an ablation study to investigate the impact of the two modules in the proposed Dual GNN framework.
Specifically, we consider two variants of the Dual learning framework:
(1) {\em Primary+Cluster}. For this variant, we drop the auxiliary module
but keep the primary module and the fine-grained spectral clustering loss.
That is, we perform training with an objective of $\mathcal{L}_{CE} + \mathcal{L}_{sc}$.
(2) {\em Auxiliary+Cluster}. For this variant, we drop the primary module but keep its encoder $f_\Theta$. 
We perform training with an objective of $\mathcal{\widetilde{L}}_{CE} + \mathcal{L}_{sc}$.

We conducted experiments with GCN as the baseline model,
which can be treated as a special variant 
of the Dual method that uses only the primary module. 
The comparison results with different label rates are reported in Table~\ref{tab:ablation}.
We can see that all variants 
have performance drop from the full Dual model
and the performance degradations are substantial with small label rates. 
The variant of Primary+Cluster has similar performance as the base model. 
This suggests the fined-grained clustering loss alone cannot help the primary module,
while the auxiliary module which constructs new graph structures from fine-grained clustering 
plays an essential role in promoting message propagation across the graph and overcoming the 
local overfitting problems caused by small label rates. 
By dropping the primary module, the performance of Auxiliary+Cluster drops significantly.
This suggests the auxiliary module can be easily misled by its constructed dense adjacency matrix
without using the primary module to learn discriminative node embeddings from the original graph and labeled nodes.  
The results in Table \ref{tab:ablation} demonstrate the contribution of 
each module 
and 
the importance of the joint Dual learning framework.

\section{Conclusion}

In this paper, we proposed a novel Dual GNN learning framework to address the drawbacks of standard GNN models
on handling scarce labeled nodes and noisy graph structures
for semi-supervised node classification.
The proposed framework consists of two modules. 
The primary GNN module works on the original input graph, 
while the auxiliary module 
employs a new adjacency matrix constructed using fine-grained spectral clustering to 
facilitate messages propagation across the graph. 
The two modules and the spectral clustering are learned under a joint minimization framework.
This dual learning framework can be applied on many existing GNN baselines.
We conducted experiments with four GNN baseline models and the experimental results
demonstrated that the proposed framework 
is robust to scarce labels and noisy graph structures with missing edges,
and can significantly improve the GNN baselines on benchmark datasets. 

\bibliographystyle{abbrvnat}
\bibliography{references.bib}

\end{document}